\documentclass[10pt,twocolumn,letterpaper]{article}

\usepackage{iccv}
\usepackage{times}
\usepackage{epsfig}
\usepackage{graphicx}
\usepackage{amsmath}
\usepackage{amssymb}
\usepackage{algorithm} 
\usepackage{algpseudocode}
\usepackage{multirow}
\usepackage{cite}
\usepackage{array}

\usepackage[breaklinks=true,bookmarks=false]{hyperref}

\iccvfinalcopy 

\def\iccvPaperID{****} 

\ificcvfinal\fi

\begin{document}
\makeatletter
\newcommand{\printfnsymbol}[1]{%
  \textsuperscript{\@fnsymbol{#1}}%
}
\makeatother

\title{Deep Elastic Networks with Model Selection for Multi-Task Learning}

\author{Chanho Ahn\thanks{Indicates equal contribution}\\
Dept. of ECE and ASRI\\
Seoul National University\\
{\tt\small mychahn@snu.ac.kr}
\and
Eunwoo Kim\printfnsymbol{1}\\
Department of Engineering Science\\
University of Oxford\\
{\tt\small ekim@robots.ox.ac.uk}
\and
Songhwai Oh\\
Dept. of ECE and ASRI\\
Seoul National University\\
{\tt\small songhwai@snu.ac.kr}
}

\maketitle
\ificcvfinal\thispagestyle{empty}\fi

\newcolumntype{L}[1]{>{\raggedright\let\newline\\\arraybackslash\hspace{0pt}}m{#1}}
\newcolumntype{C}[1]{>{\centering\let\newline\\\arraybackslash\hspace{0pt}}m{#1}}
\newcolumntype{R}[1]{>{\raggedleft\let\newline\\\arraybackslash\hspace{0pt}}m{#1}}

\begin{abstract} 
    In this work, we consider the problem of instance-wise dynamic network model selection for multi-task learning.
    To this end, we propose an efficient approach to exploit a compact but accurate model in a backbone architecture for each instance of all tasks.
    The proposed method consists of an estimator and a selector.
    The estimator is based on a backbone architecture and structured hierarchically.
    It can produce multiple different network models of different configurations in a hierarchical structure.
    The selector chooses a model dynamically from a pool of candidate models given an input instance.
    The selector is a relatively small-size network consisting of a few layers, which estimates a probability distribution over the candidate models when an input instance of a task is given.
    Both estimator and selector are jointly trained in a unified learning framework in conjunction with a sampling-based learning strategy, without additional computation steps.
    We demonstrate the proposed approach for several image classification tasks compared to existing approaches performing model selection or learning multiple tasks.
    Experimental results show that our approach gives not only outstanding performance compared to other competitors but also the versatility to perform instance-wise model selection for multiple tasks.
\end{abstract}

\section{Introduction}\label{sec:intro}

\begin{figure}[t]
    \centering
    \includegraphics[scale=0.27]{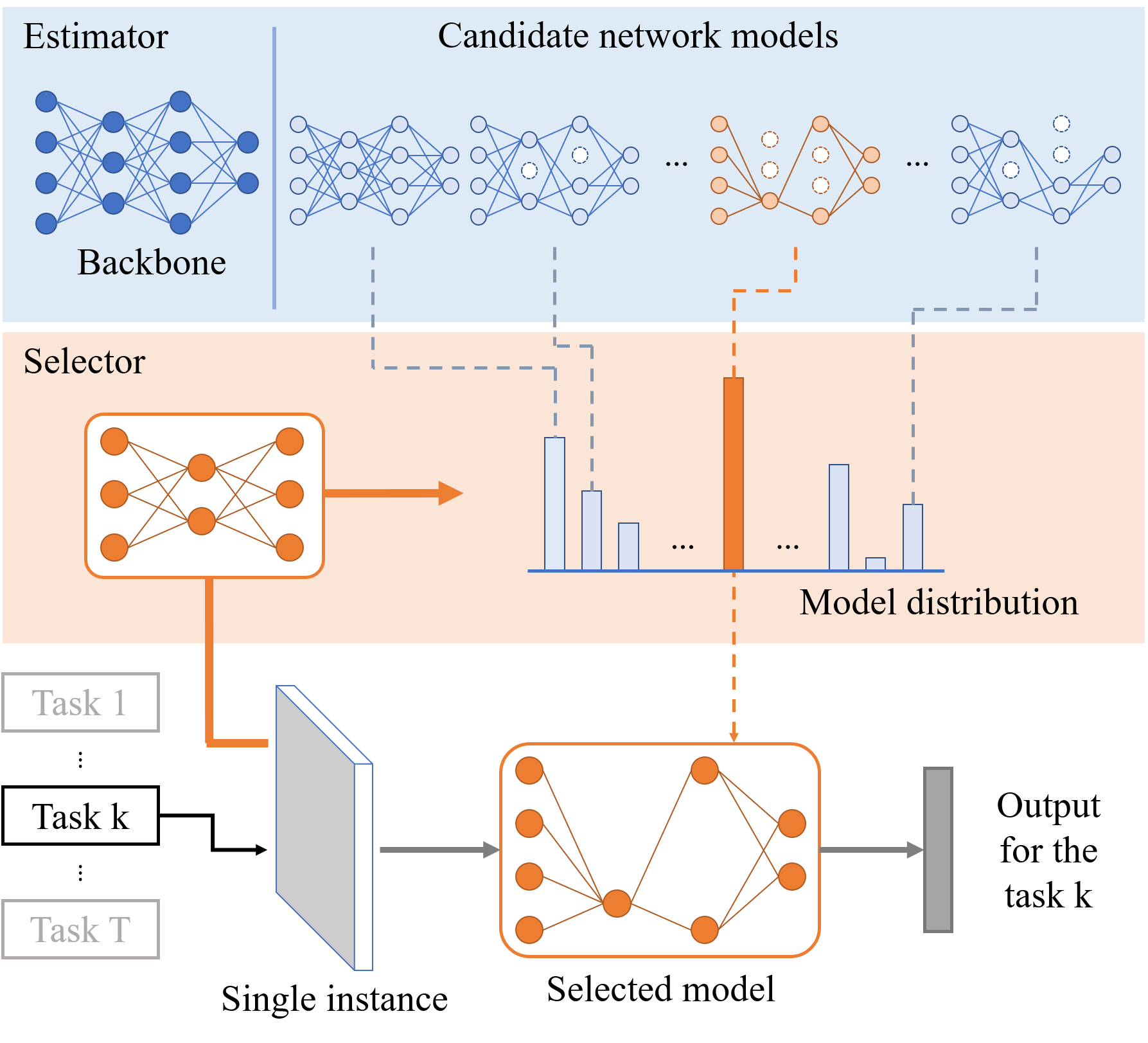}
    \caption{
    An overview of the proposed framework, which consists of an estimator and a selector.
    The estimator, whose structure is identical to the backbone network, includes multiple internal networks (models) of different configurations and scales.
    The selector outputs a probability distribution over the candidate models given an instance from a task.
    The model with the highest probability is chosen from the estimator to perform the assigned task.
    }
    \label{fig:sum}
\end{figure}

Multi-task learning (MTL) \cite{multitask} simultaneously learns multiple tasks to improve generalization performance for the tasks.
Most of recent MTL approaches \cite{packnet, cross, beyond, routing} are based on deep neural networks (DNNs) which have outstanding performance compared to traditional machine learning methods in computer vision and machine learning, such as image classification \cite{resnet, resnext}, object detection \cite{detection}, and pose estimation \cite{stacked}, to name a few.

Since it is believed that MTL methods using a DNN require a huge number of parameters and computing resources, a compact network with a small number of parameters and low computational complexity is highly desirable for many practical applications, such as mobile and embedded platforms \cite{mobile}.
To address this, there have been studies on designing compact DNNs, such as network pruning \cite{P1, S2}, knowledge distillation \cite{K1, K2}, network architecture search \cite{nas}, and adaptive model compression \cite{R1, D1, D3}.
However, these prior works have been applied to a single task problem and multiple tasks have been little considered in a single framework.

The MTL problem has a potential issue that the required number of parameters may increase depending on the number of tasks \cite{multitask}.
However, a single shared model for multiple tasks may cause performance degradation when associated tasks are less relevant \cite{routing}.
To avoid this issue, recent approaches \cite{nested, dvn} proposed a network architecture which can contain several sub-models to assign the them to multiple tasks.
Despite their attempts for MTL, they require human efforts to construct sub-models from the network architecture and assign the model to each task.
For more flexible and adaptive model assignment for multiple tasks, it is desired to realize a model selection approach which automatically determines a proper sub-model depending on a given instance.

In this work, we aim to develop an instance-aware dynamic model selection approach for a single network to learn multiple tasks.
To that end, we present an efficient learning framework that exploits a compact but high-performing model in a backbone network, depending on each instance of all tasks.
The proposed framework consists of two main components of different roles, termed an \textit{estimator} and a \textit{selector} (see Figure \ref{fig:sum}).
The estimator is based on a backbone (baseline) network, such as VGG \cite{vgg} or ResNet \cite{resnet}.
It is structured hierarchically based on modularized blocks which consist of several convolution layers in the backbone network.
It can produce multiple network models of different configurations and scales in a hierarchy.
The selector is a relatively small network compared to the estimator and outputs a probability distribution over candidate network models for a given instance.
The model with the highest probability is chosen by the selector from a pool of candidate models to perform the task.
Note that the approach is learned to choose a model corresponding to each instance throughout all tasks.
This makes it possible to share the common models or features across all tasks \cite{decaf, nested}.
We design the objective function to achieve not only competitive performance but also resource efficiency (i.e., compactness) required for each instance.
Inspired by \cite{policy}, we introduce a sampling-based learning strategy to approximate the gradient for the selector which is hard to derive exactly.
Both the estimator and the selector are trained in a unified learning framework to optimize the associated objective function, which does not require additional efforts (e.g., fine-tuning) performed in existing works \cite{D3, nas}.

We perform a number of experiments to demonstrate the competitiveness of the proposed method, including model selection and model compression problems when a single or multiple tasks are given.
For the experiments, we use an extensive set of benchmark datasets: CIFAR-10 and CIFAR-100 \cite{cifar}, Tiny-ImageNet\footnote{https://tiny-imagenet.herokuapp.com/}, STL-10 \cite{stl}, and ImageNet \cite{imagenet}.
The experimental results on different learning scenarios show that the proposed method outperforms existing state-of-the-art approaches.
Notably, our approach addresses both model selection and multi-task learning simultaneously in a single framework without introducing additional resources, making it highly efficient.

\section{Related Work} \label{sec:rw}

\noindent \textbf{Model selection.}
In order to reduce the burden of an expert for designing a compact network, architecture search methods \cite{nas} were proposed to explore the space of potential models automatically.
To shrink the daunting search space which usually requires a time-consuming exploration, methods based on a well-developed backbone structure find an efficient model architecture by compressing a given backbone network \cite{R1, R2}.
Furthermore, the recent studies realizing such strategy \cite{D1, D3, graphs} determine a different network model for each instance to reduce an additional redundancy.
However, they usually achieve the lower performance compared to their backbone network \cite{D1, graphs} or require additional fine-tuning process \cite{D3}.
In contrast to them, we propose an efficient learning framework which can achieve better performance than the backbone network due to the dynamic model search and also does not includes an additional fine-tuning stage.
Besides, our approach can be applied to learn multiple tasks simultaneously in a single framework, while aforementioned methods are limited to a single task.
\\

\noindent \textbf{Multi-task learning.}
The purpose of multi-task learning (MTL) is to develop a learning framework that jointly learns multiple tasks \cite{multitask}.
Note that we focus on a MTL method that learns a single DNN architecture for memory efficiency.
There are several recent studies \cite{zip, beyond, cross} that proposed a network structure in which parameters can be efficiently shared across tasks.
Other approaches \cite{packnet, nested, dvn} suggest a single architecture which includes multiple internal networks (or models) so that they can assign different models to multiple tasks without increasing the parameters.
However, they use a fixed model structure for each task and it requires expert efforts to assign the model to each task.
In contrast, we propose a dynamic model selection for MTL which determines a proper model automatically for a given instance.
Even if a recent MTL method \cite{routing} attempts model selection by a routing mechanism, it does not consider an optimized network structure associated with the number of parameters or FLOPs.

\begin{figure*}[t]
\centering
\includegraphics[scale = 0.31]{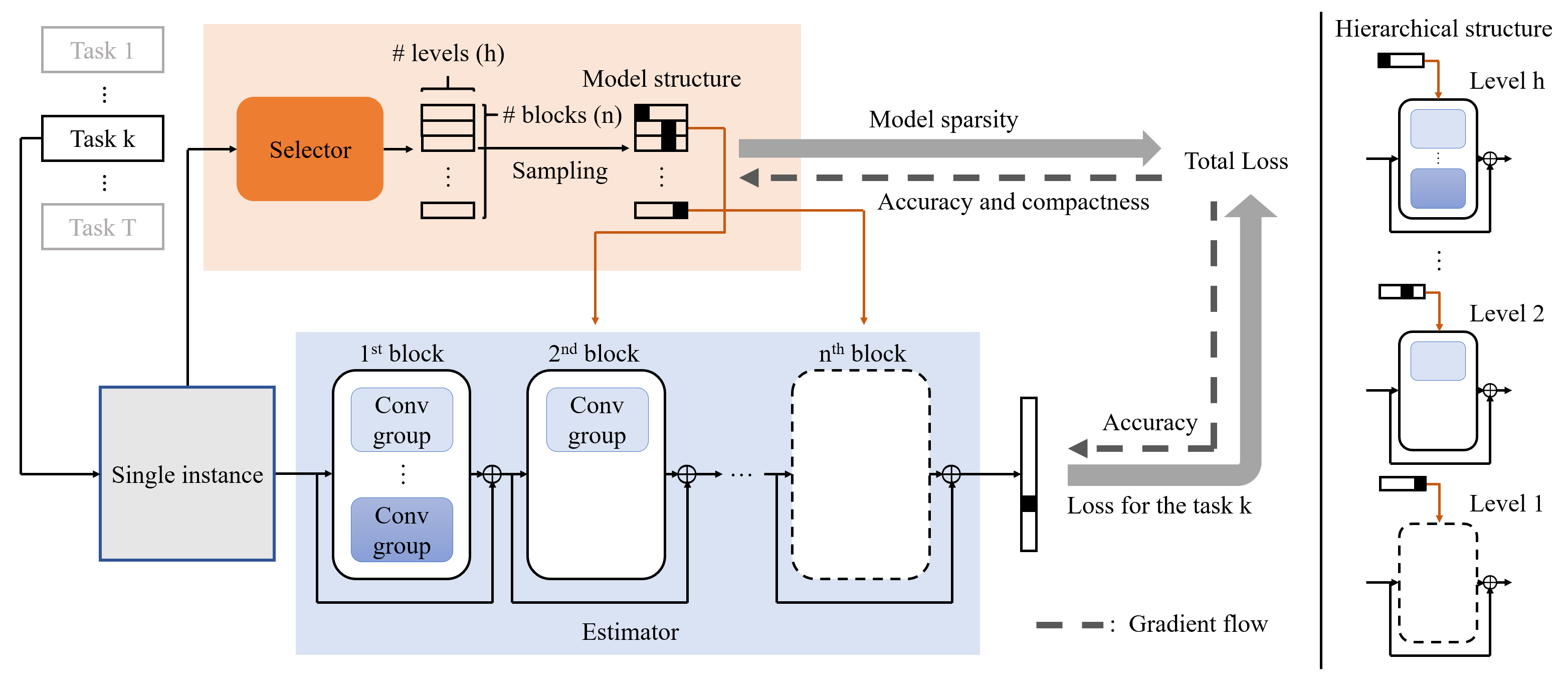}
\caption {
A graphical representation of the proposed framework which is based on a backbone network (a residual network \cite{resnet}).
The framework consists of an estimator and a selector.
The estimator, whose structure is identical to the backbone network, contains $n$ disjoint blocks.
A block is defined as a collection of consecutive convolution layers (a block is the same as a residual block while keeping the number of channels).
To simplify the hierarchical structure of each block, convolution layers in each block are divided into multiple groups.
As shown on the right side of the figure, lower levels of hierarchy contain fewer convolution groups and higher levels contains more groups.
The estimator can produce different network models by selecting convolution groups from zero to all groups in every block.
The selector outputs a probability distribution over the convolution groups in every block, and a network model is determined from the distribution.
The overall loss function consists of a prediction loss term (e.g., cross-entropy) from the determined network model and a sparse regularization term.
}
\label{fig:overall}
\end{figure*}

\section{Approach}

\subsection{Overall framework} \label{sec:net}

The goal of the proposed method is to develop a dynamic model selection framework when an input instance drawn from one of the target tasks is given.
The proposed framework consists of two different components:
an ``estimator $f$'' which is a network of the same size to the target backbone network and contains multiple different models of different network configurations, and a ``selector $g$'' which reveals a model with the highest probability in the estimator.
Both estimator and selector are constructed based on a CNN-based architecture, and the selector is designed to be much smaller than the estimator (see Section \ref{sec:exp}).
The proposed approach explores a model search space and identifies an efficient network model to perform the given task in an instance-wise manner.
The overall framework of the proposed approach is illustrated in Figure \ref{fig:sum}.

Note that there are a vast number of candidate models produced by the estimator, and this makes it difficult for the selector to explore the extensive search space.
As a simplification strategy of the daunting task, we use a \textit{block} notation to shrink the search space over the candidate models.
A block is defined as a disjoint collection of multiple convolution (or fully connected) layers.
The block is constructed as a hierarchical structure such that a lower level of hierarchy only refers fewer channels of hidden layers in the block and a higher level refers more channels, maintaining input and output dimensions of the block.
Moreover, the lowest level of hierarchy can be constructed without any channels when the block is equivalent to a residual module \cite{resnet}.
This is similar to a layer skipping method in \cite{D3}.
The hierarchical structure in a block is illustrated in Figure \ref{fig:overall}.

We determine a model structure by selecting a level of hierarchy in each block as follows: $z = (l_1, l_2, \cdots, l_n)$, where $n$ is the number of the blocks in the estimator $f$ and $l_i$ denotes the selected level in the $i$-th block.
Namely, a network model is collected in the estimator when the network model structure $z$ is given.
The inference of the determined network model is represented as follows:
\begin{equation}
f(\cdot;\theta_{est}, z, t):\mathcal{X}_t \rightarrow \mathcal{Y}_t,
\end{equation}
where $\theta_{est}$ is a set of parameters in the estimator, and $\mathcal{X}_t$ and $\mathcal{Y}_t$ denote input and output domains for task $t$, respectively.
To address different input or output dimensions, we assume that the task ID is given beforehand.

The goal of the selector $g$ is to find an appropriate network model for a given instance from a task by inferring the probability distribution over candidate models in the estimator.
As mentioned earlier, we design the selector to produce a set of probability distributions over the modularized blocks (with their levels of hierarchy) as follows:
\begin{equation}
g(\cdot; \theta_{sel}):\mathcal{X}_t \rightarrow [0, 1]^{h\times n},
\end{equation}
where $\theta_{sel}$ is a set of parameters of the selector and $h$ is the number of levels of hierarchy in each block.
We define the output of the selector as $\mathcal{C} \in [0, 1]^{h\times n}$ and each column of $\mathcal{C}$ reveals probabilities of selecting levels in the corresponding block (i.e., $\sum_i \mathcal{C}_{ij} = 1, ~ \forall j$).
Then, the probability of a candidate model for an instance $x$ can be calculated as 
\begin{equation}\label{eq:prob_m}
    \begin{split}
    P_{g(x;\theta_{sel})}\left(z;x\right) = \prod_{i=1}^n \mathcal{C}_i(l_i;x), \\
    s.t. ~~ z = (l_1, \cdots , l_n), \qquad~
    \end{split}
\end{equation}
where $\mathcal{C}_i(l_i;x) \in [0, 1]$ denotes the $l_i$-th element of the $i$-th column of $\mathcal{C}$, which means the probability that the $l_i$-th level is selected in the $i$-th block for an input $x$.
Thus, we can represent up to $h^n$ different candidate models, and one of them is selected to produce its corresponding model to perform the task.
The overall framework is shown in Figure \ref{fig:overall}.

\subsection{Optimization} \label{sec:opt}

The proposed approach is optimized to perform multi-task learning in an instance-wise manner within a single framework.
We denote a set of datasets $\mathcal{D}$ as $\mathcal{D}=\{(x,y,t)|(x,y) \in \mathcal{D}_t, ~ \forall t \}$,
where $x$ and $y$ are an image and a label, respectively, and $\mathcal{D}_t$ is a dataset for task $t$.
The proposed model selection problem is to minimize the loss functions for instances of all tasks while imposing the model size compact:
\begin{equation}\label{eq:final_eq}
    \begin{split}
    &J(\theta_{est}, \theta_{sel}) = \\
    &\mathbb{E}_{{(x,y,t)}\in\mathcal{D}, z\sim g(x; \theta_{sel})}
    \left[ \mathcal{L} ( f(x;\theta_{est}, z, t), y ) + \mathcal{S}(z) \right],
    \end{split}
\end{equation}
where $\mathcal{L}(\cdot, \cdot)$ denotes a classification loss function (e.g., cross-entropy). $\mathcal{S}(z)$ is a sparse regularization term on the model structure $z$, which is defined as:
\begin{equation}\label{eq:density}
    \mathcal{S}(z) = \rho \cdot \left(\frac{1}{n}\sum_{i=1}^n d_i(l_i)\right)^2, ~~~~ s.t. ~~ z =(l_1, \cdots, l_n),
\end{equation}
where $d_i(l_i)$ gives the ratio of the number of parameters determined by $l_i$ from the total number of parameters in the $i$-th block, and $\rho$ is a weighting factor.
The square function in (\ref{eq:density}) can help enforce high sparsity ratio, and we have empirically found that it performs better than other regularization function, such as the $l_1$-norm.

The proposed approach involves alternating optimization steps for two sets of parameters, $\theta_{est}$ and $\theta_{sel}$\footnote{ We call this alternating step as a stage.}.
While $\theta_{est}$ can be updated by a stochastic gradient descent optimizer (SGD \cite{sgd}), the gradient with respect to \eqref{eq:final_eq} for $\theta_{sel}$ is difficult to calculate without a exact expected value in \eqref{eq:final_eq}.
For this reason, we introduce a sampling-based approach to approximate the gradient.
To describe the approximation, we introduce $R$ which is equivalent to the loss function for $\theta_{sel}$ as follows:
\begin{equation}\label{eq:m}
    \begin{split}
        &J_s(\theta_{sel}) = 
        \mathbb{E}_{{(x,y,t)}\in\mathcal{D}, z\sim g(x; \theta_{sel})}
        \left[R(z; x, y, t)\right] \\
        &s.t. ~~ R(z;x, y, t) \triangleq \mathcal{L}(f(x;\theta_{est}, z, t), y) + \mathcal{S}(z).
    \end{split}
\end{equation}
Then, we can approximate the gradient value with sampled model structures, following the strategy in \cite{policy}:
\begin{equation}\label{eq:tr}
    \begin{split}
        &\nabla_{\theta_{sel}} J_s(\theta_{sel}) \\
        &= \mathbb{E}_{{(x,y,t)}\in\mathcal{D}} \left[ \sum_{\forall z} R(z;x, y, t) \nabla_{\theta_{sel}} P(z;x) \right] \\
        &= \mathbb{E}_{{(x,y,t)}\in\mathcal{D}} \left[ \sum_{\forall z} R(z;x, y, t) P(z;x) \frac{\nabla_{\theta_{sel}} P(z;x)}{P(z;x)}\right]\\
        & = \mathbb{E}_{{(x,y,t)}\in\mathcal{D}, z\sim g(x; \theta_{sel})} \left[ R(z;x, y, t) \nabla_{\theta_{sel}} \log P(z;x) \right]\\
        & \approx \mathbb{E}_{{(x,y,t)}\in\mathcal{D}} \left[\sum_{z\in \mathcal{Z}} \frac{R(z;x, y, t)}{|\mathcal{Z}|} \nabla_{\theta_{sel}} \log P(z;x) \right],
    \end{split}
    \raisetag{70pt}
\end{equation}
where $P(z;x) \triangleq P_{g(x;\theta_{sel})}(z;x)$.
The last line approximates the expectation as the average for some randomly chosen samples $z$'s which are collected from the same probability distribution when $x$ is given.
$\mathcal{Z}$ is a set of the collected $z$'s and $|\mathcal{Z}|$ denotes the number of samples in $\mathcal{Z}$.

Note that the sampling scheme follows the common strategy in the reinforcement learning literature \cite{deepRL}. 
However, this can often lead to a worse network structure when the selected model is poor \cite{exp}.
As a remedy, we apply the $\epsilon$-greedy method \cite{epsilon} to allow more dynamic exploration at the earliest training time.
In addition, we would like to note that the performance of the selected model may be sensitive to the initial distribution of the selector.
For this reason, we use the following pre-determined distributions of the network model in the initial stage:
\begin{equation} \label{eq:initdist}
    p(z_i) = 
    \begin{cases}
     (1-\tau)/h^n + \tau, &\mbox{if $z_i = z^*$,} \\
     (1-\tau)/h^n, &\mbox{otherwise,}
    \end{cases}
\end{equation}
where $\tau$ is a weighting factor, $p(z_i)$ is a probability that the model structure $z_i$ is selected, and $z^*$ denotes the full model structure which includes all parameters in the estimator.
In this work, we set $\tau$ to $0.75$ in all conducted experiments.
We increase the probability that the full model structure is selected more often in the initial stage, and it shows better performance compared to other initial distributions, such as a uniform distribution.

The overall training procedure of the proposed method, named deep elastic network (DEN), is summarized in Algorithm \ref{alg:whole}, where S denotes the number of stages.
We optimize two sets of parameters, $\theta_{est}$ and $\theta_{sel}$, during the several stages of the training process.
At each stage, one of the above parameter sets is trained until it reaches the local optima.

\begin{algorithm}
    \caption{Deep Elastic Network (DEN)}\label{alg:whole}
    \begin{algorithmic}[1]
        \State \textbf{Input:} $\mathcal{D}$, $\rho$
        \State \textbf{Initialize:} $\theta_{est}$, $\theta_{sel}$ $\leftarrow$ Xavier-initializer \cite{xavier}, S
        \State $p$ $\leftarrow$ initial distribution in \eqref{eq:initdist}
        \For{s = 1 to S}
            \Repeat
                \State derive a model structure from $p$
                \State update $\theta_{est}$ w.r.t. \eqref{eq:final_eq}
            \Until{convergence}
            \State decay the learning rate for $\theta_{est}$
            \Repeat
                \State update $\theta_{sel}$ using the gradient \eqref{eq:tr}
            \Until{convergence}
            \State decay the learning rate for $\theta_{sel}$
            \State $p$ $\leftarrow$ $g(\cdot;\theta_{sel})$
        \EndFor
    \end{algorithmic}
\end{algorithm}

\section{Experiments} \label{sec:exp}

\subsection{Experimental setup}\label{setup}

\noindent \textbf{Datasets.}
We evaluated the proposed framework on several classification datasets as listed in Table \ref{tab:dataset}.
For CIFAR-10, CIFAR-100, Tiny-ImageNet, and STL-10 datasets, we used the original image size.
Mini-ImageNet is a subset of ImageNet \cite{imagenet} which has 50 class labels and each class has 800 training instances.
We resized each image in the Mini-ImageNet dataset to $256\times 256$ and center-cropped it to have the size of $224 \times 224$.
As pre-processing techniques, we performed the random horizontal flip for all datasets and added zero padding of four pixels before cropping for CIFAR, Tiny-ImageNet, and STL-10 datasets.
CIFAR-100 dataset includes two types of class categories for each image: 20 coarse and 100 fine classes.
We used both of them for hierarchical classification; otherwise, we used the fine classes for the rest of the experiments.
\\

\noindent \textbf{Scenarios.}
We evaluated three scenarios for multi-task learning (MTL) and one scenario for network compression.
For MTL, we organized two scenarios (\textit{M1}, \textit{M2}) using multiple datasets and one scenario (\textit{M3}) using a single dataset with hierarchical class categories.
For the first scenario, \textit{M1}, we used three datasets of different image scales: CIFAR-100 (32$\times$32), Tiny-ImageNet (64$\times$64), and STL-10 (96$\times$96).
For \textit{M2}, 50 labels are randomly chosen from the 1000 class labels in the ImageNet dataset and the chosen labels are separated into 10 disjoint subsets (tasks) each of which has 5 labels.
\textit{M3} is a special case of MTL (we call it hierarchical classification), which aims to predict two different labels (coarse and fine classes) simultaneously for each image.
CIFAR-100 was used for the scenario \textit{M3}.
We also conducted the network compression scenario (\textit{C1}) as a single task learning problem for CIFAR-10 and CIFAR-100, respectively.
\\

\begin{table}[t]
\caption{Summary of the datasets. 
The size represents the width and height of an input image for each dataset.
\# train and \# test denote the number of images in the train and test sets, respectively.} \label{tab:dataset}
    \footnotesize
    \centering
    \begin{tabular}{|l||c|c|c|c|}
        \hline
         {Dataset} & Size & \# {train} & \# {test} & \# {classes} \\
         \hline \hline
         CIFAR-10 \cite{cifar} & 32  & 50,000 & 10,000 & 10\\
         CIFAR-100 \cite{cifar} & 32  & 50,000 & 10,000 & 100\\
         Tiny-ImageNet & 64 & 100,000 & 10,000 & 200\\
         STL-10 \cite{cifar} & 96  & 5,000 & 8,000 & 10\\
         Mini-ImageNet \cite{routing} &224 & 40,000 & 2,500 & 50\\
         \hline
    \end{tabular}
\end{table}

\begin{table*}[t]
\caption{Accuracy ($\%$) on three tasks (datasets) of different input scales based on two different backbone networks: (a) ResNet-42 \cite{resnet} and (b) WRN-32-4 \cite{wide}. 
We also provide FLOPs and the number of parameters for all compared methods.
[$\cdot$] denotes the number of required network models to perform the same tasks.
Baseline requires three models to perform different tasks.
$\rho$ controls sparsity of the proposed method in (\ref{eq:density}).
The bold and underline letters represent the best and the second best accuracy, respectively.
}\label{tab:m1r}
\centering
\footnotesize
\begin{tabular}{|l||C{2.0cm}|C{2.0cm}|C{2.0cm}|C{2.0cm}|C{2.0cm}|}
     \hline
     Dataset & Baseline \cite{resnet} & NestedNet \cite{nested} & PackNet* \cite{packnet} &  DEN ($\rho = 1$) & DEN ($\rho = 0.1$) \\
     \hline \hline
     CIFAR-100 (32$\times$32) & \underline{75.05} &  74.53 & 72.22 & 74.30 & \textbf{75.11}\\
     Tiny-ImageNet (64$\times$64) & \underline{57.22} & 56.71 & 56.49 & 56.74 & \textbf{60.21}\\
     STL-10 (96$\times$96) & 76.25 & 82.54 & 80.78 & \underline{83.90} & \textbf{87.58}\\
     \hline 
     Average & 69.51 & 71.26 & 69.83 & \underline{71.65} & \textbf{74.30}\\
     \hline \hline
     FLOPs  & 2.91G & 1.70G & 1.70G & 1.35G & 1.61G\\
     No. parameters & 89.4M [3] & 29.8M [1] & 29.8M [1] & 29.8M [1] & 29.8M [1] \\
     \hline
     \multicolumn{6}{c}{(a) ResNet-42} \\
     \hline
     Dataset & Baseline \cite{wide} & NestedNet \cite{nested} & PackNet* \cite{packnet} &  DEN ($\rho = 1$)  & DEN ($\rho = 0.1$)\\
     \hline \hline
     CIFAR-100 (32$\times$32) & 75.01  &  74.09 & 73.56 & \underline{75.43} & \textbf{75.65} \\
     Tiny-ImageNet (64$\times$64)  & \textbf{58.89} & 57.87 & 57.17 & 58.17 & \underline{58.25} \\
     STL-10 (96$\times$96) & 79.88 & 83.78 & 84.15 & \underline{87.54} & \textbf{87.56}\\
     \hline 
     Average  & 71.26  & 71.91 & 71.63 & \underline{73.71} & \textbf{73.82}\\
     \hline \hline
     FLOPs & 2.13G  & 1.24G & 1.24G & 1.13G & 1.14G\\
     No. parameters & 22.0M [3] & 7.35M [1] & 7.35M [1] & 7.35M [1]  & 7.35M [1] \\
     \hline
     \multicolumn{6}{c}{(b) WRN-32-4}
\end{tabular}
\end{table*}

\noindent \textbf{Implementation details.}
We used ResNet-$l$ \cite{resnet} and WRN-$l$-$r$ \cite{wide} as backbone networks in the MTL scenarios, where $l$ is the number of layers and $r$ is the scale factor on the number of convolutional channels.
We borrowed a residual network architecture designed for ImageNet \cite{imagenet} to handle large-scale images and a WRN architecture designed for CIFAR \cite{cifar} to handle small-scale images.
We also used SimpleConvNet introduced in \cite{routing, fewshot} as a backbone network for Mini-ImageNet.
SimpleConvNet consists of four 3x3 convolutional layers (32 filters) and three fully connected layers (128 dimensions for hidden units).
In the network compression scenario, we used ResNeXt-$l$ ($c \times s$d) \cite{resnext} and VGG-$l$ \cite{vgg} to apply our methods in various backbones, where $c$ and $s$d are the number of individual convolution blocks and unit depth of the convolution blocks in each layer, respectively \cite{resnext}.
The backbone networks are used as baseline methods performing an individual task in each scenario.
To build the structure of the estimator, we defined a block as a residual module \cite{resnet} and as two consecutive convolution layers for VGG networks.
Then we split each block into multiple convolution groups along the channel dimension (2 or 3 groups in our experiments) to construct a hierarchical structure.
Note that the lowest level of hierarchy does not have any convolution groups for ResNet, WRN, and ResNeXt, but has one group for VGG, and SimpleConvNet.
The selector was designed with a network which is smaller than the estimator.
The size of the selector is stated in each experiment.

For the proposed method, named deep elastic network (DEN), the estimator was trained by the SGD optimizer with Nesterov momentum of 0.9, with the batch size of 256 for large-scale dataset (ImageNet) and 128 for other datasets.
The ADAM optimizer \cite{adam} was used to learn the selector with the same batch size.
The initial learning rates were 0.1 for the estimator and 0.00001 for the selector, and we decayed the learning rate with a factor of 10 when it converges (three or four decays happened in all experiments for both estimator and selector).
All experiments are conducted in the TensorFlow environment \cite{tensorflow}.
\\

\noindent \textbf{Compared methods.}
We compared with four state-of-the-art algorithms considering resource efficiency for multi-task learning: PackNet*, NestedNet \cite{nested}, Routing \cite{routing}, and Cross-stitch \cite{cross}.
PackNet* is a variant of PackNet \cite{packnet}, which considers group-wise compression along the channel dimension, in order to achieve practical speed-up like ours.
Both PackNet* and NestedNet divide convolutional channels into multiple disjoint groups and construct a hierarchical structure such that the $i$-th level of hierarchy includes $i$ divided groups (the number of levels of hierarchy corresponds to the number of tasks).
When updating the $i$-th level of hierarchy, NestedNet considers parameters in the $i$-th level but PackNet* considers parameters except those in the ($i$-$1$)-th level.
For Routing and Cross-stitch, we used the results in \cite{routing} under the same circumstance.
We also compared with BlockDrop \cite{D3}, N2N \cite{R1}, Pruning (which we termed) \cite{P3}, and NestedNet \cite{nested} for the network compression problem.
Note that we reported FLOPs and the number of parameters of the proposed method for the estimator in all experiments.

\subsection{Multi-task learning} \label{sec:multi}

For the first scenario \textit{M1} (on three tasks), we used both ResNet-42 and WRN-32-4 as backbone networks, respectively.
The three tasks, Tiny-ImageNet, CIFAR-100, and STL-10, are assigned to the levels of hierarchy for PackNet* and NestedNet from the lowest to highest levels, respectively.
The number of parameters and FLOPs of the selector are 1.49M and 0.15G for the ResNet-42 backbone and 0.37M and 0.11G for the WRN-32-4 backbone, respectively.
The baseline method requires three separate networks, each trained independently.
Table \ref{tab:m1r} shows the results with respect to accuracy, FLOPs and the number of parameters of the compared methods.
Here, FLOPs denotes the average FLOPs for multiple tasks, and the number of parameters is measured from all networks required to perform the tasks.
Overall, our approach outperforms other methods including the baseline method.
In addition, we provide the results by varying the weighting factor $\rho$ of our sparse regularizer in \eqref{eq:density}.
As shown in the table, the performance is better when $\rho$ is lower, and more compact model is selected when $\rho$ is higher.

For the scenario \textit{M2}, SimpleConvNet was used as a backbone network.
Since the scenario contains a larger number of tasks than the previous scenario, PackNet* and NestedNet, which divide the model by human design, cannot be applied.
We divided the convolution parts which takes most of the FLOPs in the network into two levels such that lowest level of hierarchy contains half the parameters of the highest level.
The number of parameters of the selector is 0.4M, whereas the number of parameters of the estimator is 0.8M.
In this scenario, the selector is not much smaller than the estimator because the estimator is constructed in sufficiently small size.
However, the number of parameters of the selector for other scenarios are negligible compared to those of the estimator.
The accuracy, FLOPs and the number of parameters of the compared methods are reported in Table \ref{tab:m2r}.
The result of the compared methods are reported in the work in \cite{routing}.
For a fair comparison, we experimented with our algorithm on two input scales.
The proposed method shows a significant performance improvement compared to the other methods, even though ours uses lower average FLOPs than others for evaluations.
In addition, the proposed method has similar FLOPs to the comparison methods even when dealing with large scale inputs and has outstanding performance.
Note that since the number of parameters and FLOPs are not precisely reported in the paper, we provide lower bounds.

\subsection{Hierarchical classification}

For the scenario \textit{M3}, we dealt with CIFAR-100 which has coarse and fine class categories for each image as described in Section \ref{setup}.
WRN-32-4 was used as a backbone network for this scenario.
We compared with PackNet* and NestedNet, and the lowest and highest levels of hierarchy for them were allocated to perform the coarse and fine classifications, respectively.
The structure of the selector in our method is equal to that in the scenario \textit{M1}.

\begin{table}[t]
\caption{Accuracy ($\%$) on the Mini-ImageNet dataset, FLOPs, and the number of parameters for all compared methods.
Baseline uses the different last fully-connected layer for different tasks and shares other layers across the tasks.
The bold and underline letters represent the best and the second best accuracy, respectively.
DEN$_{s}$ denotes results for input images of size 84$\times$84.
}\label{tab:m2r}
\centering
\footnotesize
\begin{tabular}{|C{2.0cm}||C{1.4cm}|C{1.4cm}|C{1.4cm}|}
     \hline
     Method & Accuracy & FLOPs & No. params \\
     \hline \hline
     Baseline & 51.03 & 27.4M & 0.14M\\
     Cross-stitch \cite{cross} & 56.03 & $>$ 27.4& $>$ 0.14M \\
     Routing \cite{routing} & 58.97 & 27.4M & $>$ 0.14M \\
     DEN$_{s}$ ($\rho = 1$) & 60.78 & 18.7M & 0.14M \\
     DEN$_{s}$ ($\rho = 0.1$) & 62.62 & 19.4M & 0.14M \\
     \hline
     DEN ($\rho = 1$) & \underline{63.20} & 33.3M & 0.85M \\
     DEN ($\rho = 0.1$) & \textbf{65.23} & 39.1M & 0.85M \\
     \hline
\end{tabular}
\end{table}

Table \ref{tab:m3r} summarizes the results of the compared methods for the coarse and fine classification problems.
Our approach shows the best accuracy while giving the lowest FLOPs compared to the competitors except the baseline method for both problems.
Furthermore, the proposed method has higher performance than the baseline method on average.
Since each image has two different tasks (coarse and fine class categories), the selector exploits the same model structure and thus gives almost the same FLOPs.

\subsection{Network compression}

The goal of the network compression problem is to design a compact network model from a given backbone network while minimizing the performance degradation.
We applied the proposed method to the network compression problem which is a single-task learning problem.
We compared with BlockDrop \cite{D3} and NestedNet \cite{nested} on two backbone networks: ResNeXt \cite{resnext} and VGG \cite{vgg}.
Since BlockDrop is developed for residual networks, we compared with it using ResNeXt.
The CIFAR-10 and CIFAR-100 \cite{cifar} datasets were used for the scenario, respectively.
To verify the efficiency of the proposed method, we constructed our method with four levels of hierarchy for ResNeXt-29 (8$\times$64d) and three levels for ResNeXt-29 (4$\times$64d), respectively.
The numbers of parameters of the selector are 3.9M and 3.6M for VGG and ResNeXt backbone networks, respectively.

Table \ref{tab:cifar} summarizes the classification accuracy of the compared approaches for the backbone networks.
Overall, the proposed method shows the highest accuracy compared to other compression approaches.
Our results with different $\rho$ show that $\rho$ can provide a trade-off between the network size and its corresponding accuracy.
We also tested the proposed method (estimator) with a random selector, which reveals a model structure randomly among the candidate models in the estimator, to compare it with our model selection method.
From the result, we can observe that the accuracy of the random selector is lower than the proposed selector, which reveals that the selector has potential to explore the desired model.
Moreover, we compared with the state-of-the-art network compression methods, N2N \cite{R1}, and Pruning \cite{P3}, whose results were reported from their papers \cite{R1, P3}.
Our approach has 94.47\% classification accuracy using 5.8M parameters and the Pruning method has 94.15\% accuracy using 6.4M parameters on the CIFAR-10 dataset.
The proposed method also shows better performance than N2N and Pruning methods on the CIFAR-100 dataset.

\begin{table}[t]
\footnotesize
    \centering
    \caption{Hierarchical classification results on CIFAR-100.
    Baseline (WRN-32-4\cite{wide}) requires two models to perform different tasks.
    The bold and underline letters represent the best and the second best accuracy, respectively.
    }
    \label{tab:m3r}
    \begin{tabular}{|c||C{1.5cm}|C{1.5cm}|C{1.5cm}|}
        \hline
        Method & Accuracy & FLOPs & No. params \\
        \hline \hline
        Baseline \cite{wide} & 83.53 & 2.91G & 14.7M\\
        NestedNet \cite{nested} & \underline{84.55} & 1.46G & 7.35M\\
        PackNet* \cite{packnet} & 84.53 & 1.46G & 7.35M \\
        DEN ($\rho = 1$) & \textbf{84.87} & 1.37G & 7.35M\\
        \hline
        \multicolumn{4}{c}{(a) Coarse classification (20)} \\
        \hline
        Method & Accuracy & FLOPs & No. params \\
        \hline \hline
        Baseline \cite{wide} & \textbf{76.32} & 2.91G & 14.7M \\
        NestedNet \cite{nested} & 75.84 & 2.91G & 7.35M\\
        PackNet* \cite{packnet} & 75.65 & 2.91G & 7.35M\\
        DEN ($\rho = 1$) & \underline{75.93} & 1.37G & 7.35M\\
        \hline
        \multicolumn{4}{c}{(b) Fine classification (100)}
    \end{tabular}
\end{table}

\begin{table*}[ht]
\footnotesize
\caption{Network compression results on the CIFAR dataset.
For FLOPs, we refer to the compression ratio from the baseline network for each model and dataset.
The bold and underline letters represent the best and the second best accuracy, respectively.
``rand sel'' denotes that the random model structure is used without using the selector in the proposed method.
The results of NestedNet are obtained from the lowest (L) to the highest (H) levels of hierarchy (including the intermediate level (M) for ResNeXt-29 (8 $\times$ 64d)).
}
\centering
\label{tab:cifar}
    \begin{tabular}{|c|c||C{1.4cm}|C{1.4cm}|C{1.4cm}||C{1.4cm}|C{1.4cm}|C{1.4cm}|}
        \hline
        \multicolumn{2}{|c||}{Dataset}&\multicolumn{3}{c||}{CIFAR-10} & \multicolumn{3}{c|}{CIFAR-100}\\
        \hline 
        Backbone & Method & Acc (\%) & No. params & FLOPs & Acc (\%) & No. params & FLOPs \\
        \hline \hline
        \multirow{4}{*}{VGG-16}
        & Baseline \cite{vgg} & \textbf{92.52} & 38.9M & 1.0$\times$ & \textbf{69.64} & 38.9M & 1.0$\times$ \\
        & NestedNet \cite{nested}, L & 91.29 & 19.4M & 2.0$\times$ & 68.10 & 19.4M & 2.0$\times$\\
        & NestedNet \cite{nested}, H & \underline{92.40} & 38.9M & 1.0$\times$ & \underline{69.01} & 38.9M & 1.0$\times$ \\
        & DEN ($\rho=0.1$) & {92.31} & 18.5M & 2.4$\times$ & {68.87} & 18.9M & 1.7$\times$ \\
        \hline\hline
        ResNet-18 & \multirow{2}{*}{N2N \cite{R1}} & 91.97 & 2.12M & $-$ & 69.64 & 4.76M & $-$\\
        ResNet-34 &  & 93.54 & 3.87M & $-$ & 70.11 & 4.25M & $-$\\ \hline
        \multirow{2}{*}{ResNet-50} & Pruning \cite{P3} & 94.15 & 6.44M & $-$ & 74.10 & 9.24M & $-$\\
        &DEN ($\rho=1$) & 94.50 & 4.25M & $-$ & 77.98 & 4.67M & $-$\\
        \hline\hline
        \multirow{8}{*}{ResNeXt-29 (8 $\times$ 64d)}
        & Baseline \cite{resnext} & \textbf{94.61} & 22.4M & 1.0$\times$ & \textbf{78.73} & 22.4M & 1.0$\times$\\
        & NestedNet \cite{nested}, L & 93.56 & 5.6M & 4.0$\times$ & 74.83 & 5.6M & 4.0$\times$\\
        & NestedNet \cite{nested}, M & 93.64 & 11.2M & 2.0$\times$ & 74.98 & 11.2M & 2.0$\times$\\
        & NestedNet \cite{nested}, H & {94.13} & 22.4M & 1.0$\times$ & 76.16 & 22.4M & 1.0$\times$\\
        & BlockDrop \cite{D3} & 93.56 & 16.9M & 1.2$\times$ & 78.35 & 15.5M & 1.4$\times$\\
        & DEN $+$ rand sel & 90.55 & 9.8M & 2.3$\times$ & 69.67 & 9.8M & 2.3$\times$\\
        & DEN ($\rho=1$) & 91.45 & 4.1M & 5.5$\times$ & {78.27} & 7.3M & 3.0$\times$\\
        & DEN ($\rho=0.1$) & \textbf{94.61} & 8.7M & 2.7$\times$ & \underline{78.68} & 13.5M & 1.9$\times$\\
        \hline \hline
        \multirow{7}{*}{ResNeXt-29 (4 $\times$ 64d)}
        & Baseline \cite{resnext} & \underline{94.37} & 11.2M & 1.0$\times$ & \textbf{77.95} & 11.2M & 1.0$\times$\\
        & NestedNet \cite{nested}, L & 93.59 & 5.6M & 2.0$\times$ & 75.70 &  5.6M & 2.0$\times$\\
        & NestedNet \cite{nested}, H & {94.11} & 11.2M & 1.0$\times$ & 76.36 & 11.2M & 1.0$\times$\\
        & BlockDrop \cite{D3} & 93.07 & 6.53M & 1.7$\times$ & {77.23} & 7.47M & 1.5$\times$\\
        & DEN (rand sel) & 87.33 & 5.6M & 2.0$\times$ & 65.44 & 5.6M & 2.0$\times$\\
        & DEN ($\rho=1$) & 93.38 & 5.4M & 2.1$\times$ & 76.71 & 5.6M & 2.0$\times$\\
        & DEN ($\rho=0.1$) & \textbf{94.47} & 5.8M & 1.9$\times$ & \underline{77.58} & 6.3M & 1.8$\times$\\
        \hline \hline
        ResNeXt-29 (2 $\times$ 64d) & Baseline \cite{resnext} & 93.60 & 5.6M & $-$ & 76.54 & 5.6M & $-$\\
        \hline
    \end{tabular}
\end{table*}

\subsection{Quantitative results} \label{sec:quan}

\begin{figure*}[t]
    \centering
    \includegraphics[scale=0.5]{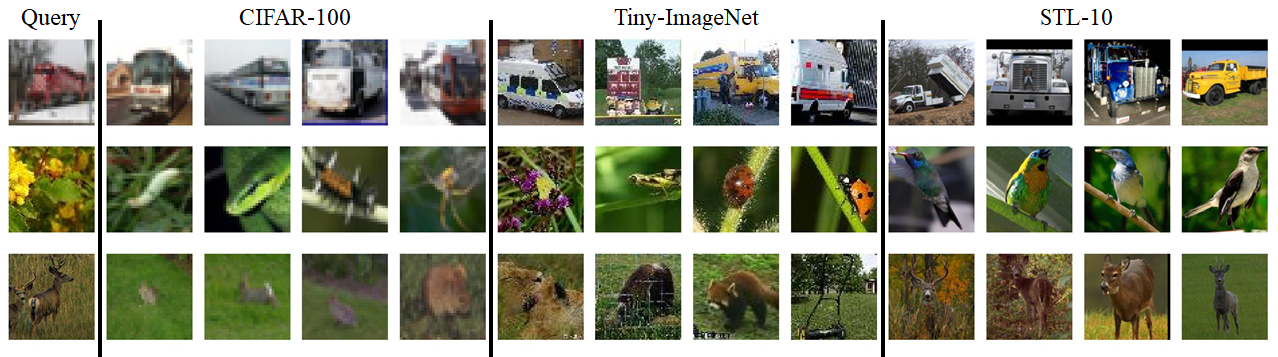}
    \caption{
    Sampled images from each task (dataset) which have the similar model distribution to that of the query images (first column).
    The query images belong to CIFAR-100, Tiny-ImageNet, and STL-10 from top to bottom, respectively.
    }
    \label{fig:pca}
\end{figure*}

The proposed instance-wise model selection for multi-task learning can associate similar features for similar images, which means that similar model structures can be selected for similar images.
To verify this, we chose one input image (query) at each task and derived its output model distribution from the selector.
Here, we measured the similarity between the distributions using $l_2$-distance.
Then we collected four samples from each task, whose corresponding outputs have the similar model distribution to the query image.
To do so, we constructed the proposed method based on the WRN-32-4 backbone architecture for the three tasks (datasets): CIFAR-100, Tiny-ImageNet, and STL-10.
We set the size of input image to $32\times 32$ for all the datasets to see the similarity under the same image scale.
Figure \ref{fig:pca} shows some selected images from all tasks for each query image.
The results show that instance-wise model selection can be a promising strategy for multi-task learning as it can reveal the common knowledge across the tasks.
We provide model distributions for instances from the test set in supplementary materials, along with the ablation study of using different numbers of levels.

\section{Conclusion}
In this work, we have proposed an efficient learning approach to perform resource-aware dynamic model selection for multi-task learning.
The proposed method contains two main components of different roles, an estimator which produces multiple candidate models, and a selector which exploits a compact and competitive model among the candidate models to perform the designated task.
We have also introduced a sampling-based optimization strategy to address the discrete action space of the potential candidate models.
The proposed approach is learned in a single framework without introducing many additional parameters and much training efforts.
The proposed approach has been evaluated on several problems including multi-task learning and network compression.
The results have shown the outstanding performance of the proposed method compared to other competitors.

\noindent \textbf{Acknowledgements:} This research was supported in part
by Institute of Information \& Communications Technology Planning \& Evaluation (IITP) grant funded by the Korea government (MSIT) No. 2019-0-01190, [SW Star Lab] Robot Learning: Efficient, Safe, and Socially-Acceptable Machine Learning, and No. 2019-0-01371, Development of Brain-Inspired AI with Human-Like Intelligence,
and by AIR Lab (AI Research Lab) of Hyundai Motor Company through HMC-SNU AI Consortium Fund.

{\small
\bibliographystyle{ieee_fullname}
\bibliography{egbib}

\begin{thebibliography}{10}\itemsep=-1pt

\bibitem{tensorflow}
Mart{\'\i}n Abadi, Paul Barham, Jianmin Chen, Zhifeng Chen, Andy Davis, Jeffrey
  Dean, Matthieu Devin, Sanjay Ghemawat, Geoffrey Irving, Michael Isard, et~al.
\newblock Tensorflow: A system for large-scale machine learning.
\newblock In {\em 12th Symposium on Operating Systems Design and Implementation
  (OSDI 16)}, 2016.

\bibitem{R2}
Karim Ahmed and Lorenzo Torresani.
\newblock {MaskConnect}: Connectivity learning by gradient descent.
\newblock In {\em European Conference on Computer Vision (ECCV)}. Springer,
  2018.

\bibitem{R1}
Anubhav Ashok, Nicholas Rhinehart, Fares Beainy, and Kris~M Kitani.
\newblock N2{N} learning: network to network compression via policy gradient
  reinforcement learning.
\newblock {\em arXiv preprint arXiv:1709.06030}, 2017.

\bibitem{sgd}
L{\'e}on Bottou.
\newblock Large-scale machine learning with stochastic gradient descent.
\newblock In {\em Proceedings of COMPSTAT'2010}, pages 177--186. Springer,
  2010.

\bibitem{multitask}
Rich Caruana.
\newblock Multitask learning.
\newblock {\em Machine learning}, 28(1):41--75, 1997.

\bibitem{stl}
Adam Coates, Andrew Ng, and Honglak Lee.
\newblock An analysis of single-layer networks in unsupervised feature
  learning.
\newblock In {\em Proceedings of International Conference on Artificial
  Intelligence and Statistics}, 2011.

\bibitem{decaf}
Jeff Donahue, Yangqing Jia, Oriol Vinyals, Judy Hoffman, Ning Zhang, Eric
  Tzeng, and Trevor Darrell.
\newblock {DeCAF}: A deep convolutional activation feature for generic visual
  recognition.
\newblock In {\em International Conference on Machine Learning (ICML)}, pages
  647--655, 2014.

\bibitem{xavier}
Xavier Glorot and Yoshua Bengio.
\newblock Understanding the difficulty of training deep feedforward neural
  networks.
\newblock In {\em Proceedings of International Conference on Artificial
  Intelligence and Statistics}, 2010.

\bibitem{P1}
Song Han, Jeff Pool, John Tran, and William Dally.
\newblock Learning both weights and connections for efficient neural network.
\newblock In {\em Advances in Neural Information Processing Systems (NIPS)},
  2015.

\bibitem{resnet}
Kaiming He, Xiangyu Zhang, Shaoqing Ren, and Jian Sun.
\newblock Deep residual learning for image recognition.
\newblock In {\em Proceedings of the IEEE conference on Computer Vision and
  Pattern Recognition (CVPR)}, 2016.

\bibitem{zip}
Xiaoxi He, Zimu Zhou, and Lothar Thiele.
\newblock Multi-task zipping via layer-wise neuron sharing.
\newblock In {\em Advances in Neural Information Processing Systems (NIPS)},
  2018.

\bibitem{K1}
Geoffrey Hinton, Oriol Vinyals, and Jeff Dean.
\newblock Distilling the knowledge in a neural network.
\newblock {\em arXiv preprint arXiv:1503.02531}, 2015.

\bibitem{mobile}
Andrew~G Howard, Menglong Zhu, Bo Chen, Dmitry Kalenichenko, Weijun Wang,
  Tobias Weyand, Marco Andreetto, and Hartwig Adam.
\newblock {MobileNets}: Efficient convolutional neural networks for mobile
  vision applications.
\newblock {\em arXiv preprint arXiv:1704.04861}, 2017.

\bibitem{P3}
Yiming Hu, Siyang Sun, Jianquan Li, Xingang Wang, and Qingyi Gu.
\newblock A novel channel pruning method for deep neural network compression.
\newblock {\em arXiv preprint arXiv:1805.11394}, 2018.

\bibitem{nested}
Eunwoo Kim, Chanho Ahn, and Songhwai Oh.
\newblock {NestedNet}: Learning nested sparse structures in deep neural
  networks.
\newblock In {\em Proceedings of the IEEE conference on Computer Vision and
  Pattern Recognition (CVPR)}, 2018.

\bibitem{dvn}
Eunwoo Kim, Chanho Ahn, Philip~HS Torr, and Songhwai Oh.
\newblock Deep virtual networks for memory efficient inference of multiple
  tasks.
\newblock In {\em Proceedings of the IEEE conference on Computer Vision and
  Pattern Recognition (CVPR)}, 2019.

\bibitem{adam}
Diederik~P Kingma and Jimmy Ba.
\newblock Adam: A method for stochastic optimization.
\newblock {\em arXiv preprint arXiv:1412.6980}, 2014.

\bibitem{cifar}
Alex Krizhevsky, Vinod Nair, and Geoffrey Hinton.
\newblock Cifar-10 and cifar-100 datasets.
\newblock {\em URL: https://www.cs.toronto.edu/kriz/cifar.html (visited on
  Mar.1, 2016)}, 2009.

\bibitem{imagenet}
Alex Krizhevsky, Ilya Sutskever, and Geoffrey~E Hinton.
\newblock Imagenet classification with deep convolutional neural networks.
\newblock In {\em Advances in Neural Information Processing Systems (NIPS)},
  2012.

\bibitem{D1}
Ji Lin, Yongming Rao, Jiwen Lu, and Jie Zhou.
\newblock Runtime neural pruning.
\newblock In {\em Advances in Neural Information Processing Systems (NIPS)},
  2017.

\bibitem{detection}
Tsung-Yi Lin, Priyal Goyal, Ross Girshick, Kaiming He, and Piotr Doll{\'a}r.
\newblock Focal loss for dense object detection.
\newblock {\em IEEE Transactions on Pattern Analysis and Machine Intelligence
  (TPAMI)}, 2018.

\bibitem{packnet}
Arun Mallya and Svetlana Lazebnik.
\newblock {PackNet}: Adding multiple tasks to a single network by iterative
  pruning.
\newblock In {\em Proceedings of the IEEE conference on Computer Vision and
  Pattern Recognition (CVPR)}, 2018.

\bibitem{beyond}
Elliot Meyerson and Risto Miikkulainen.
\newblock Beyond shared hierarchies: Deep multitask learning through soft layer
  ordering.
\newblock {\em arXiv preprint arXiv:1711.00108}, 2017.

\bibitem{cross}
Ishan Misra, Abhinav Shrivastava, Abhinav Gupta, and Martial Hebert.
\newblock Cross-stitch networks for multi-task learning.
\newblock In {\em Proceedings of the IEEE Conference on Computer Vision and
  Pattern Recognition (CVPR)}, pages 3994--4003, 2016.

\bibitem{deepRL}
Volodymyr Mnih, Koray Kavukcuoglu, David Silver, Alex Graves, Ioannis
  Antonoglou, Daan Wierstra, and Martin Riedmiller.
\newblock Playing atari with deep reinforcement learning.
\newblock {\em arXiv preprint arXiv:1312.5602}, 2013.

\bibitem{stacked}
Alejandro Newell, Kaiyu Yang, and Jia Deng.
\newblock Stacked hourglass networks for human pose estimation.
\newblock In {\em European Conference on Computer Vision (ECCV)}. Springer,
  2016.

\bibitem{fewshot}
Sachin Ravi and Hugo Larochelle.
\newblock Optimization as a model for few-shot learning.
\newblock 2017.

\bibitem{K2}
Adriana Romero, Nicolas Ballas, Samira~Ebrahimi Kahou, Antoine Chassang, Carlo
  Gatta, and Yoshua Bengio.
\newblock {FitNets}: Hints for thin deep nets.
\newblock {\em arXiv preprint arXiv:1412.6550}, 2014.

\bibitem{routing}
Clemens Rosenbaum, Tim Klinger, and Matthew Riemer.
\newblock Routing networks: Adaptive selection of non-linear functions for
  multi-task learning.
\newblock {\em arXiv preprint arXiv:1711.01239}, 2017.

\bibitem{vgg}
Karen Simonyan and Andrew Zisserman.
\newblock Very deep convolutional networks for large-scale image recognition.
\newblock {\em arXiv preprint arXiv:1409.1556}, 2014.

\bibitem{policy}
Richard~S Sutton, David~A McAllester, Satinder~P Singh, and Yishay Mansour.
\newblock Policy gradient methods for reinforcement learning with function
  approximation.
\newblock In {\em Advances in Neural Information Processing Systems (NIPS)},
  2000.

\bibitem{epsilon}
Michel Tokic.
\newblock Adaptive $\varepsilon$-greedy exploration in reinforcement learning
  based on value differences.
\newblock In {\em Annual Conference on Artificial Intelligence}, pages
  203--210. Springer, 2010.

\bibitem{graphs}
Andreas Veit and Serge Belongie.
\newblock Convolutional networks with adaptive inference graphs.
\newblock In {\em European Conference on Computer Vision (ECCV)}. Springer,
  2018.

\bibitem{S2}
Wei Wen, Chunpeng Wu, Yandan Wang, Yiran Chen, and Hai Li.
\newblock Learning structured sparsity in deep neural networks.
\newblock In {\em Advances in Neural Information Processing Systems (NIPS)},
  2016.

\bibitem{D3}
Zuxuan Wu, Tushar Nagarajan, Abhishek Kumar, Steven Rennie, Larry~S Davis,
  Kristen Grauman, and Rogerio Feris.
\newblock {BlockDrop}: Dynamic inference paths in residual networks.
\newblock In {\em Proceedings of the IEEE conference on Computer Vision and
  Pattern Recognition (CVPR)}, 2018.

\bibitem{exp}
Jeremy Wyatt.
\newblock Exploration and inference in learning from reinforcement.
\newblock 1998.

\bibitem{resnext}
Saining Xie, Ross Girshick, Piotr Doll{\'a}r, Zhuowen Tu, and Kaiming He.
\newblock Aggregated residual transformations for deep neural networks.
\newblock In {\em Proceedings of the IEEE conference on Computer Vision and
  Pattern Recognition (CVPR)}, 2017.

\bibitem{wide}
Sergey Zagoruyko and Nikos Komodakis.
\newblock Wide residual networks.
\newblock {\em arXiv preprint arXiv:1605.07146}, 2016.

\bibitem{nas}
Barret Zoph and Quoc~V Le.
\newblock Neural architecture search with reinforcement learning.
\newblock {\em arXiv preprint arXiv:1611.01578}, 2016.

\end{thebibliography}
}

\newpage

\clearpage

\appendix

\section{Appendix}

\subsection{Details of Hierarchical Structure}

\begin{figure}[h]
    \centering
    \includegraphics[scale=0.3]{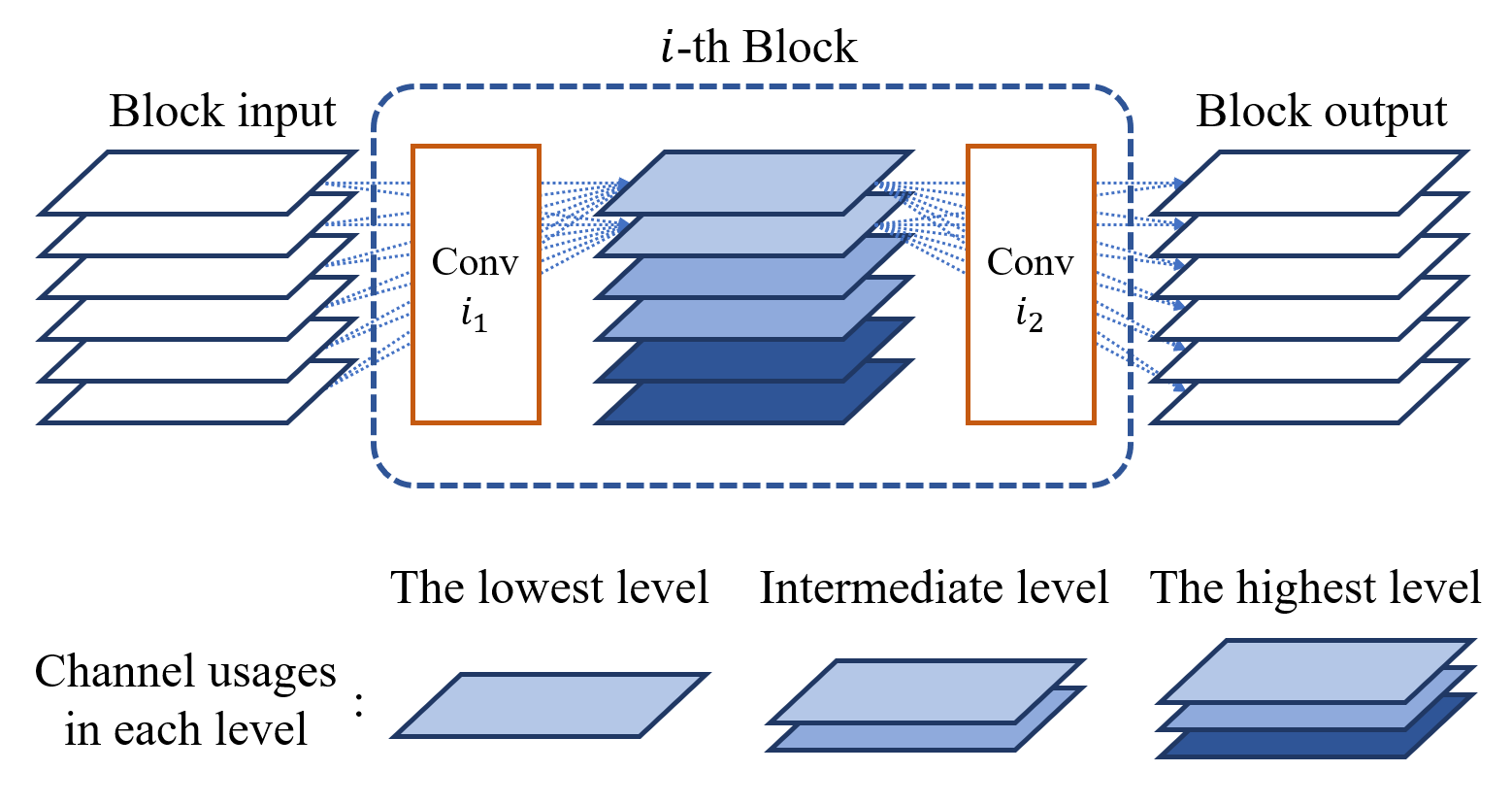}
    \caption{
    An example of the hierarchical structure in a block (the lowest level of hierarchy is shown as by dotted lines).
    Each hierarchical structure of a block contains different numbers of channels in the layer such that the lower level of hierarchy uses less channels and higher level uses more channels.
    The number of convolution filters used at each level depends on the channel usage.
    }
    \label{fig:block}
\end{figure}

The estimator of the proposed method can produce multiple network models of different sizes based on the hierarchical structure in a block.
To control the actual speed-up for inference, each hierarchy accesses a different number of channels in each convolution layer.
The ratio of the required number of channels for each level can be adjusted.
As shown in Figure \ref{fig:block}, the lowest level of hierarchy is represented and it accesses only a few channels.
The highest level of hierarchy contains all channels in the figure.
If the block is based on a residual block \cite{resnet}, the lowest level does not include any channels.

\subsection{Ablation Studies}

We evaluated the performance depending on the number of levels in each block or depending on the initial model distribution.
We used WRN-32-4 \cite{wide} as a backbone network and the CIFAR-100 dataset \cite{cifar}.

\begin{figure}[h]
    \centering
    \includegraphics[scale=0.6]{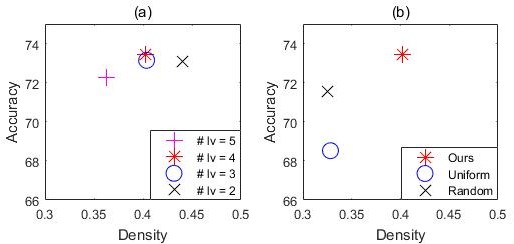}
    \caption{
    Two ablation studies: (a) performance on varying numbers of levels, and (b) performance with different model distributions.
    ``\# lv'' is the number of levels in each block.
    ``Uniform'' and ``Random'' denote that the corresponding methods learn the estimator with a uniform model distribution and the random model distribution from the untrained selector in an initial stage, respectively.
    }
    \label{fig:abl}
\end{figure}

First, we tested the performance on varying numbers of levels.
The number of candidate models increases greatly as the number of levels increases, while the size of the selector is held fixed (the number of candidate models is $h^n$, where $h$ and $n$ are the number of levels and blocks, respectively).
Figure \ref{fig:abl} shows that the larger the number of levels, the smaller the network size can be found as exploring a larger model space.
The performance also improved incrementally until the number of levels is four.
However, when the number of levels is five, the performance is degraded due to the failure on dealing with a number of candidate models.

Second, we verified the effect of the initial model distribution.
We applied two other distributions to compare with the proposed model distribution as described in Section 3.2 in the main paper: uniform distribution (Uniform) and random distribution which is obtained from the untrained initial selector (Random).
The initial model distribution was used for training the estimator in the initial stage.
As observed in Figure \ref{fig:abl}, we can verify that learning the network with the proposed initial distribution shows the best performance.
Using the other distributions in the initial stage, the accuracy of the initial stage converged to the 2 to 3 \% lower value compared to our method.
Our approach reveals high performance in the initial stage and this affects the overall performance in Figure \ref{fig:abl}-(b).

\subsection{Model Distribution for Test Set} \label{appendix:Model}

\begin{figure}[h]
    \centering
    \includegraphics[scale=0.5]{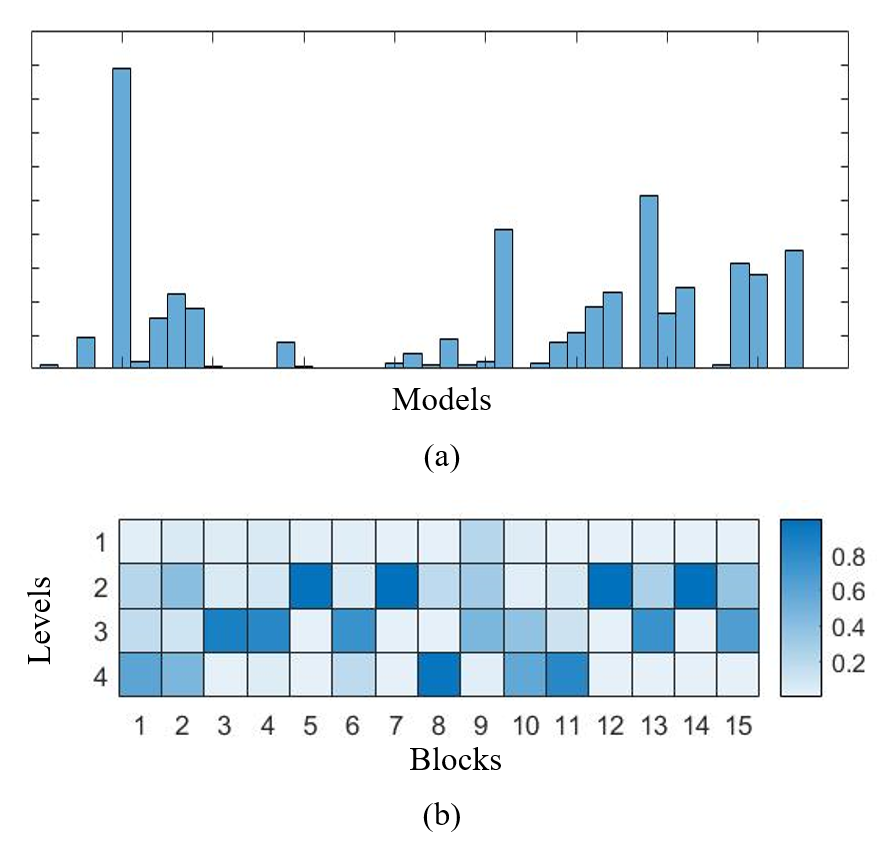}
    \caption{
    Evaluation of selected models using a network with 15 blocks and four hierarchical levels.
    (a) A histogram of models selected by the proposed algorithm on the test set.
    (b) The mean of probabilities that each level is selected by the proposed algorithm on the test set.
    }
    \label{fig:hist}
\end{figure}

We describe the model distribution for the test set to verify that diverse models can be selected depending on given input instances.
The proposed framework was trained on three datasets, CIFAR-100 \cite{cifar}, Tiny-ImageNet, and STL-10 \cite{stl}, based on a backbone network, WRN-32-4 \cite{wide}.
We designed the estimator to have 15 blocks each of which contains four levels of hierarchy.
Figure \ref{fig:hist} shows the histogram of different models which are used for instances in the test sets.
We can observe variability of selected models and the distribution of chosen models is neither deterministic nor uniform.
We also calculated the average of probabilities that each level is selected over the test set.
As shown in Figure \ref{fig:hist}-(b), the high values represent that the corresponding levels of hierarchy are frequently selected over the test set and there are common filters which are used for the most instances.

\begin{figure*}[t]
    \centering
    \includegraphics[scale=0.5]{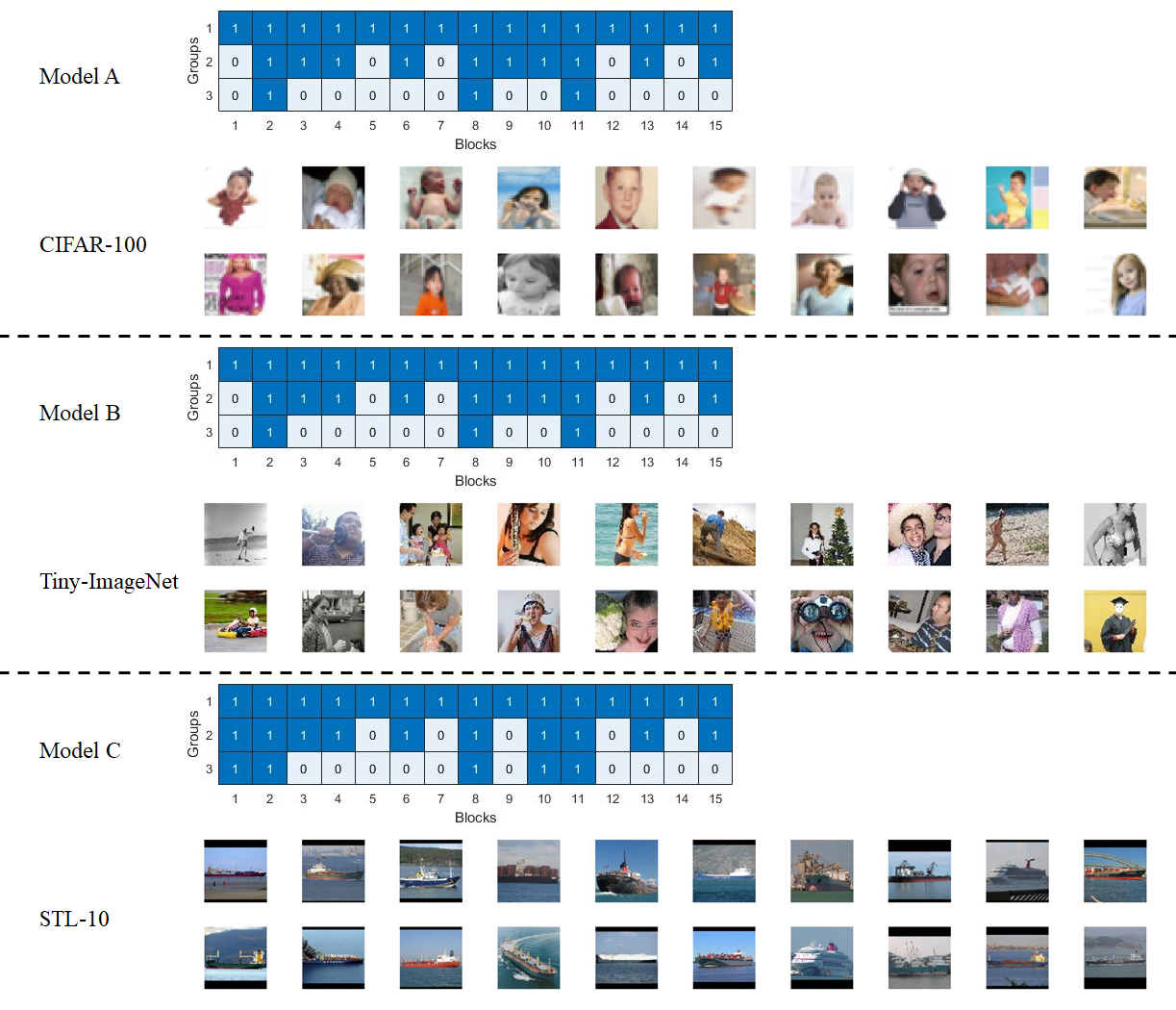}
    \caption{
    The most selected model structure for each dataset.
    The result is from the proposed framework jointly trained with three datasets: CIFAR-100, Tiny-ImageNet, and STL-10.
    ``Model'' represents convolution groups chosen by the proposed selector.
    Below the model, examples of input images which selected the model are shown.
    }
    \label{fig:mod}
\end{figure*}

From the experiment, we have found that different models are selected by different groups of images.
Examples of selected models and corresponding input images are shown in Figure \ref{fig:mod}.
Three example models are shown in the figure: Model A, B, and C.
Model A is selected for images with children and Model B is selected for images with people doing different activities.
Note that Model A and B shares the same network architecture.
Model C is selected for vessels.
Similar groups are selected for Model A and Model B while the selected groups for Model C are different from Model A and B.
We can see that each group is learned for specific features and the proposed selector explores appropriate groups for efficient inference.

\end{document}